\documentclass[lettersize,journal]{IEEEtran}
\usepackage{amsmath,amsfonts}
\usepackage{algorithmic}
\usepackage{algorithm}
\usepackage{array}
\usepackage[caption=false,font=normalsize,labelfont=sf,textfont=sf]{subfig}
\usepackage{textcomp}
\usepackage{stfloats}
\usepackage{url}
\usepackage{verbatim}
\usepackage{graphicx}
\usepackage{natbib}
\usepackage{hyperref}
\usepackage{xcolor}
\emergencystretch=\maxdimen
\setcitestyle{numbers,square}
\title{EdgeRegNet: Edge Feature-based Multimodal Registration Network between Images and LiDAR Point Clouds}
\author{Yuanchao Yue, Hui Yuan, Senior Member, IEEE, Qinglong Miao, Xiaolong Mao, Raouf Hamzaoui, Senior Member, IEEE and Peter Eisert, Senior Member, IEEE
\thanks{This work was supported in part by the National Natural Science Foundation of China under Grants 62222110 and 62172259,  the High-end Foreign Experts Recruitment Plan of Chinese Ministry of Science and Technology under Grant G2023150003L, the Taishan Scholar Project of Shandong Province (tsqn202103001), the Natural Science Foundation of Shandong Province under Grant ZR2022ZD38.(\textit{Corresponding author: Hui Yuan.})

Yuanchao Yue, Hui Yuan and Qinglong Miao are with the School of Control Science and Engineering, Shandong University, Jinan, Shandong, China (e-mail: 202234945@mail.sdu.edu.cn, huiyuan@sdu.edu.cn, william.q.miao@gmail.com).

Xiaolong Mao is with the School of software, Shandong University, Jinan, Shandong, China (e-mail: xiaolongmao@mail.sdu.edu.cn).

Raouf Hamzaoui is with the School of Engineering and Sustainable Development, De Montfort University, LE1 9BH Leicester, U.K. (e-mail: rhamzaoui@dmu.ac.uk )

Peter Eisert is with the Institut für Informatik , Humboldt-Universität zu Berlin, Germany. (e-mail: peter.eisert@hhi.fraunhofer.de)
}}
\date{May 2024}
\begin{document}

\maketitle
\begin{abstract}
Cross-modal data registration has long been a critical task in computer vision, with extensive applications in autonomous driving and robotics. Accurate and robust registration methods are essential for aligning data from different modalities, forming the foundation for multimodal sensor data fusion and enhancing perception systems' accuracy and reliability. The registration task between 2D images captured by cameras and 3D point clouds captured by Light Detection and Ranging (LiDAR) sensors is usually treated as a visual pose estimation problem. High-dimensional feature similarities from different modalities are leveraged to identify pixel-point correspondences, followed by pose estimation techniques using least squares methods. However, existing approaches often resort to downsampling the original point cloud and image data due to computational constraints, inevitably leading to a loss in precision. Additionally, high-dimensional features extracted using different feature extractors from various modalities require specific techniques to mitigate cross-modal differences for effective matching.
To address these challenges, we propose a method that uses edge information from the original point clouds and images for cross-modal registration. We retain crucial information from the original data by extracting edge points and pixels, enhancing registration accuracy while maintaining computational efficiency. The use of edge points and edge pixels allows us to introduce an attention-based feature exchange block to eliminate cross-modal disparities. Furthermore, we incorporate an optimal matching layer to improve correspondence identification. We validate the accuracy of our method on the KITTI and nuScenes datasets, demonstrating its state-of-the-art performance. Our code is publicly available on GitHub at \href{https://github.com/ESRSchao/EdgeRegNet}{https://github.com/ESRSchao/EdgeRegNet}.
\end{abstract}
\section{Introduction}
Cross-modal registration is critical in various fields such as autonomous driving systems \cite{jia2011new}\cite{an2024survey}, augmented reality \cite{466720}, mixed reality \cite{Chen2019}, and robotics \cite{8633982}. In these systems, data from different sensors exhibit diverse characteristics and errors, making registration crucial for accurately aligning and integrating this data to achieve multimodal fusion perception \cite{dong2022superfusion}. By doing so, the accuracy and reliability of the perception system are significantly enhanced. Additionally, registration helps to mitigate biased or erroneous data generated by different sources, enabling the system to perceive and understand the data more accurately. Integrating various data types through precise registration can improve the overall performance and robustness of these systems, thereby promoting more reliable and accurate decision-making processes. In the context of autonomous driving, cross-modal registration is particularly essential. It enhances vehicle environmental perception, ensuring safety and autonomy in complex road and environmental conditions. The system can build a comprehensive and reliable representation of the surroundings by accurately aligning data from LiDAR, cameras, radar, and other sensors. This is vital for making informed driving decisions and enhancing overall vehicle safety.

The registration problem between 2D images and 3D point clouds can be understood as a visual pose estimation problem. The typical steps involve finding correspondences between 2D pixels and 3D points, and then using techniques such as Efficient Perspective-n-Point (EPnP) \cite{lepetit2009ep} to estimate the transformation matrix frame-by-frame. The cross-modal registration method is different from calibration methods, including offline calibration methods \cite{kim2019extrinsic} and online calibration methods \cite{schneider2017regnet}\cite{lv2021lccnet}. Offline calibration usually uses calibration boards or manually set markers to find prominent features, such as corners, line segments, or planes to find 2D-3D correspondences, this process is usually conducted only once before the system is operational. After the calibration process has been finished, some online calibration methods are introduced to correct minor errors that appear during system operation, which are designed to adjust only small deviations that occur after initial calibration and fail when the offsets are slightly larger. However, the complex calibration process and additional algorithms make the initial preparations for data fusion time-consuming and inefficient. If the relative positions of the sensors need to be adjusted, these procedures must be repeated. 

To overcome these drawbacks, cross-modal registration methods \cite{li2021deepi2p}\cite{Ren_2023}\cite{zhou2023differentiable} have emerged to align data in different modals directly. Recent cross-modal registration techniques use specific deep learning methods to reduce cross-modal discrepancies, enabling frame-by-frame registration of point clouds and images. However, these approaches often sacrifice computational efficiency to address cross-dimensional differences, and the resulting registration accuracy has not reached a satisfactory average level. In addition, due to limited computational resources, these methods typically downsample images and point clouds and then compare the similarity between pixels of the downsampled images and points of the downsampled point clouds to establish image-point cloud correspondences, this downsample approach inevitably loses original information from the point cloud and the image during the downsampling stage, thus reducing registration accuracy. 

Registration based on key points can easily keep original information. However, for the sparse point clouds generated by real-time LiDAR scanning, extracting significant key points for matching is much more challenging due to the lack of spatial information and the difference between the dimensional \cite{an2024esc}. In our study, we found that using edge information from both images and point clouds not only effectively integrates the critical information from the raw data but also leverages the inherent saliency of edges to reduce the impact of outliers, thereby improving accuracy. Although some proposed methods \cite{10160910}\cite{wang2023p2o} have already used edge information to get a better performance in calibration tasks. However, their work primarily focuses on correcting small misalignments, where edge information from different modalities is already aligned roughly. Currently, no research has explored using these edge features directly for global cross-modal matching. 

Based on this finding, we propose a method for multimodal registration based on 2D-3D global edge features. By extracting edge points from  2D images and 3D point clouds and using a deep learning feature extractor to extract features from the edge points and pixels, the features of each modality are extracted. Then, an attention-based feature exchange block is used to update the features of different dimensions. Finally, registration is achieved through the correspondence of edge features. This approach retains the accurate original information of the image and the point cloud while minimizing computational overhead during the matching process. 
In summary, our main contributions are as follows.\\
\begin{enumerate}
    \item  We propose a cross-modal registration framework that uses edge features for the global cross-modal registration between LiDAR point clouds and camera images.
    \item  To address the inherent differences between 2D images and 3D point cloud data, we introduce a custom solution designed specifically for extracting edge pixels from images and edge points from point clouds. We also propose a method for extracting features based on these edge pixels and edge points.
    \item  We propose an attention-based feature exchange module to exchange features between 2D edge pixels and 3D edge points. This module minimizes the disparities between 2D and 3D cross-modal features, enhancing the accuracy and robustness of the registration.
    \item  We propose a loss function specifically tailored to optimize the network architecture.
\end{enumerate}

The structure of the article is as follows. Section II reviews unimodal and cross-modal registration methods. Section III introduces the problem statement. Section IV presents the network architecture of our method. Section V gives comprehensive experimental results and describes the training process. Finally, Section VI concludes the paper.
\section{RELATED WORK}
\subsection{Image Registration}
Traditional image registration methods \cite{izquierdo2003efficient}\cite{lou2014image} typically rely on feature points and their corresponding descriptors to find pixel correspondences and estimate the homography matrix between the images \cite{lee2013accurate}. Common traditional image descriptors include SIFT \cite{790410}, SURF \cite{bay2006surf}. However, these descriptors may face limitations in diverse real-world scenarios, leading to the emergence of deep learning-based methods.

SuperPoint \cite{detone2018superpoint} uses CNNs to extract feature points and descriptors from images, achieving excellent performance. Building on SuperPoint, SuperGlue \cite{sarlin2020superglue} introduces a GNN to further enhance the robustness of image matching tasks. These works demonstrate the effectiveness of using attention mechanisms and solving the optimal transport problem \cite{MAL-073} with the Sinkhorn algorithm \cite{Sinkhorn1967ConcerningNM} in matching tasks.

LightGlue \cite{lindenberger2023lightglue} optimizes SuperGlue by making the network lighter and designing a more efficient correspondence prediction method, further improving network performance. Additionally, GlueStick \cite{pautrat2023gluestick} uses edge information from images and incorporates it into the GNN network, enhancing the network's performance in complex textured scenes. We also have RCVS \cite{xie2024rcvs} to handle the video frame registration problem.
\subsection{Point Cloud Registration}
Point cloud registration tasks are typically divided into fine registration and coarse registration. For fine registration tasks, a set of point clouds with relatively small deviations is given, and algorithms are used to precisely align them. The algorithm most commonly used for this purpose is the Iterative Closest Point (ICP) \cite{121791} algorithm. Since the ICP algorithm requires an initial guess and can easily get stuck in local optima, several improved algorithms have been proposed, such as the Normal Distributions Transform (NDT) algorithm \cite{biber2003normal} and ICP-based improvements like GO-ICP \cite{6751291}.

For coarse registration tasks, the goal is to roughly align two point clouds without an initial transformation matrix.  After coarse registration, ICP can be used to optimize the result and achieve accurate alignment \cite{huang2017coarse}. Coarse registration typically relies on finding matching point pairs and then using the Perspective-n-Point (PnP) algorithm \cite{lepetit2009ep}\cite{zhang20233d}\cite{yang2023mutual} to estimate the transformation matrix \cite{yang2018aligning}\cite{an2024ol}. For extracting matching point pairs, in addition to basic methods PointNet++ \cite{qi2017pointnet++}, D3Feat \cite{9157624} uses KPConv \cite{9010002} to extract feature points from point clouds by scoring effectively. PREDATOR \cite{9577334} introduces a GNN in the process of extracting 3D feature points, thereby enhancing the network's performance in low-overlap point cloud registration tasks. IGReg  \cite{xu2024igreg} makes good use of the fusion between images and point clouds to achieve higher point cloud registration accuracy.\\ Wu \textit{et al.} \cite{wu2023accelerating} accelerate point cloud registration with low overlap using graphs and sparse convolutions. Zhao\textit{ et al.} \cite{zhao2023registration} used a learning feature-based overlap confidence prediction method to establish reliable connections to solve the overlap region uncertainty problem.
\begin{figure*}[htbp] 
    \centering
    \includegraphics[width=1\linewidth]{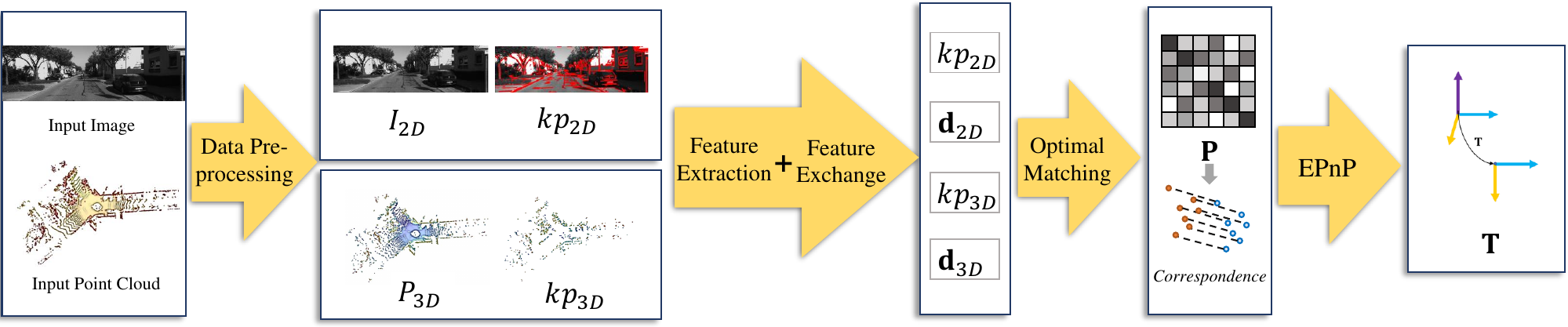}
    \caption{Overall flowchart of our method. In the data pre-processing stage, we obtain the pre-processed point cloud \(P_{3D}\) and the image \(I_{2D}\), as well as the edge points \(kp_{3D}\) of the point cloud and the edges pixels \(kp_{2D}\) of the image. After feature extraction, we acquire edge features \(\mathbf{d}_{3D}\) and \(\mathbf{d}_{2D}\). In the optimal matching process, correspondences between 2D and 3D are found using the partially assigned matrix \(\mathbf{P}\). Finally, the transformation matrix \(\mathbf{T}\) is estimated using EPnP. }
    \label{f1}
\end{figure*}
\subsection{Cross-Modality Registration}
Compared to image registration and point cloud registration tasks, which operate within the same modality, cross-modal registration tasks are more challenging. Many methods have been proposed for 2D-3D cross-modal registration. One approach involves cross-domain descriptors. Both 2D3D-MatchNet \cite{8794415} and LCD \cite{pham2019lcd} have proposed learning-based cross-domain descriptors. P2-Net \cite{Wang_2021_ICCV} efficiently extracts per-point descriptors and identifies key points in a single forward pass.

For cross-modal camera pose estimation tasks in large-scale point cloud scenes, Go-Match \cite{zhou2022geometry}, inspired by SuperGlue, uses GNN to process and match different cross-modal descriptors. Its excellent performance demonstrates the effectiveness of GNN in cross-modal networks. 

Aligning LiDAR point clouds and images is different from online calibration tasks, such as methods like LCCNet \cite{lv2021lccnet} and RGKCNet \cite{9623545} typically handle scenarios with small deviations between point clouds and images, making them unsuitable for large-scale global registration. 

Current methods for LiDAR point cloud and image registration include DeepI2P \cite{li2021deepi2p}, which uses a feature-free approach to determine whether a point cloud is within the camera's field of view, optimizing the registration through back-projection. CorrI2P \cite{Ren_2023} uses neural networks to extract individual features from images and point clouds and uses a custom feature exchange network to reduce cross-modal differences. VP2P-Match \cite{NEURIPS2023_a0a53fef} introduces a voxel branch to the existing cross-modal registration baseline, leveraging sparse convolution and image convolutional neural networks to achieve significant cross-domain matching.

These learning-based methods inevitably involve downsampling in neural networks, resulting in the loss of original data. Moreover, they can incur substantial computational costs to overcome cross-modal differences. To address these issues, we propose EdgeRegNet, which significantly improves registration accuracy while ensuring network performance.\\
\section{Problem Statement}
\begin{figure}
\centering
\includegraphics[width=1\linewidth]{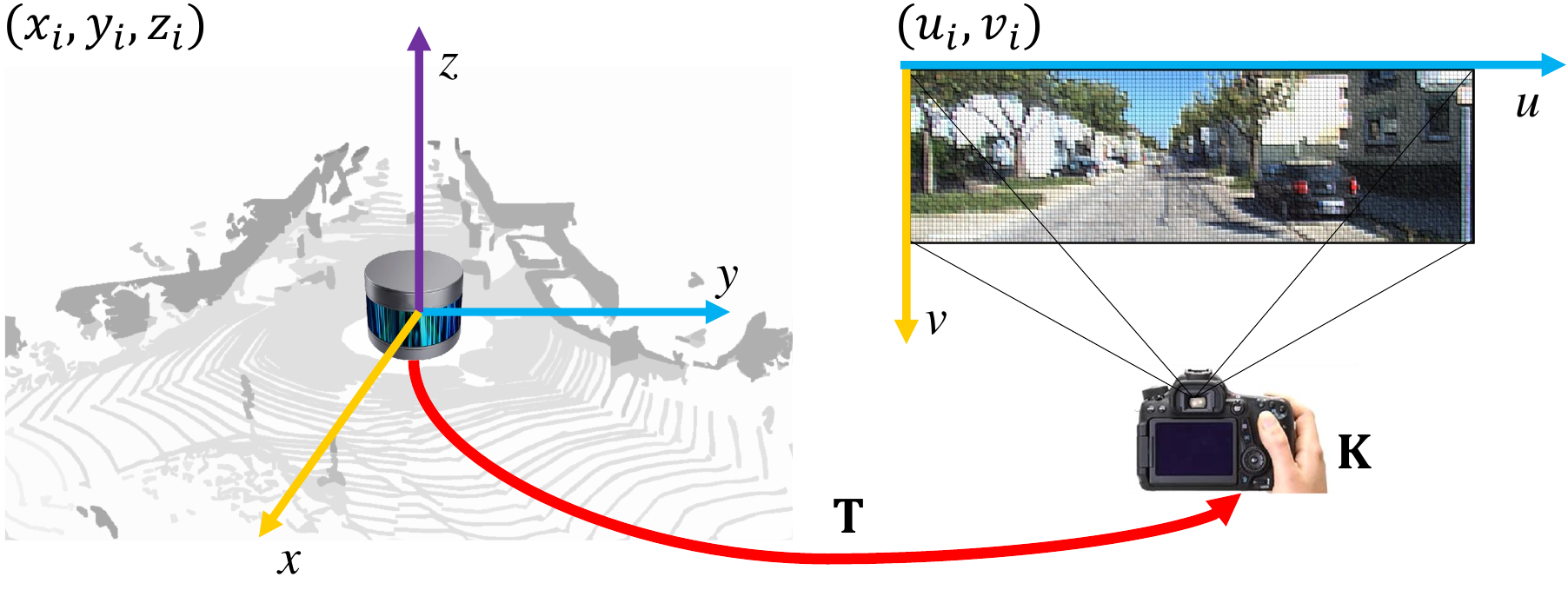} 
\caption{The left side of the figure shows the point cloud generated by the LiDAR scan and the spatial coordinate system of the point cloud data. The right side shows the 2D pixels in the camera coordinate system. \( \mathbf{T} \) represents a transformation matrix in \(\text{SE}(3)\), and \( \mathbf{K} \) denotes the camera's intrinsic parameters.} \label{f2}
\end{figure}
Given an RGB image \( \mathbf{I}_C \in \mathbb{R}^{H \times W \times 3} \) and point cloud \(\mathbf{P}_L \in \mathbb{R}^{N \times 4} \) (the 4 dimensions of the point cloud are the spatial coordinates of x,y,z, and the reflectance) we aim to find the correspondence between 2D edge pixels in the image and 3D edge points in the point cloud.

Specifically, we consider a set of 2D pixel points \( \{(u_i, v_i)\}_{i=1}^n \) in the image and a set of 3D points \( \{(x_j, y_j, z_j)\}_{j=1}^m \) in the point cloud. The objective of this task is to find a set of matching pairs \(\mathcal{M}\), \(\mathcal{M} = \{(i, j) \mid i \in \{1, 2, \ldots, n\}, j \in \{1, 2, \ldots, m\}\} \), indicating that the 2D point \((u_i, v_i)\) corresponds to the 3D point \((x_j, y_j, z_j)\). This correspondence is based on the similarity of their multichannel features \(\mathbf{F}_{2D}\) and \(\mathbf{F}_{3D}\). 

After determining \( \mathcal{M} \), we aim to estimate the transformation matrix \(\mathbf{T}\in \text{SE}(3)\) that aligns the 3D points to the 2D image coordinates using the PnP algorithm. Matrix \(\mathbf{T}\) is given by a translation vector \(\mathbf{t} \in \mathbb{R}^3\) and a rotation matrix \(\mathbf{R} \in \text{SO}(3)\) as follows:

\[
\mathbf{T} = \begin{bmatrix}
\mathbf{R} & \mathbf{t} \\
\mathbf{0} & 1
\end{bmatrix} \in \text{SE}(3).
\tag{1}
\]

Given the intrinsic matrix \(\mathbf{K}\in\mathbb{R}^{3\times 3}\) and a scaling factor \(\lambda\), we  obtain \(\mathbf{T}\) by minimizing the reprojection error:
\[
\min_{\mathbf{T}} \sum_{(i,j) \in \mathcal{M}} \left\| \begin{bmatrix} u_i \\ v_i \end{bmatrix} - \begin{bmatrix} u'_j \\ v'_j \end{bmatrix} \right\|_2,\tag{2}
\]where \((u'_j, v'_j\)) is obtained by projecting the 3D key point \((x_j,y_j, z_j)\) onto the camera plane as follows (Fig. 2)

\[
\lambda \begin{bmatrix} u'_j \\ v'_j \\ 1 \end{bmatrix} = \mathbf{K} \mathbf{T} \begin{bmatrix} x_j \\ y_j \\ z_j \\ 1 \end{bmatrix}.\tag{3}
\]

However, the accuracy of this estimate depends on finding precise correspondences \( \mathcal{M}\). Therefore, our algorithm focuses on designing a robust and accurate correspondence framework, resulting in a hybrid registration algorithm.

\section{PROPOSED METHOD}
Our network structure is illustrated in Figure \ref{f1}. Taking a 2D grayscale image and a 3D point cloud as input, EdgeRegNet first applies data pre-processing to generate point cloud and image data suitable for neural network processing (Section \ref{4A}), along with extracted 2D edge pixels and 3D edge points. Subsequently, we use the feature extraction module to extract features from the 2D edge pixels and 3D edge points (Section \ref{4B}). After obtaining these features, we update them through a feature exchange module to reduce cross-modal discrepancies, ultimately outputting the updated features(Section \ref{4C}). Next, an optimal matching module finds correspondences between 2D pixels and 3D points using a partially assigned matrix \(\mathbf{P}\) (Section \ref{4D}). Finally, the transformation matrix \(\mathbf{T}\) is obtained by applying EPnP with RANSAC \cite{10.1145/358669.358692} to the correspondences (Section \ref{4D}).
\subsection{Data Pre-processing\label{4A}}
During data pre-processing, various operations are applied to the point cloud and image to prepare them for input into the neural network. In addition to typical operations such as downsampling and normalization, extra attention is paid in our work to extract 2D edge pixels \(kp_{2D}\) and 3D edge points \(kp_{3D}\).
Mainstream cross-modal registration methods rely on similarity matching of multichannel features from downsampled point clouds and images, which limits registration accuracy due to data loss. VP2P-Match \cite{NEURIPS2023_a0a53fef} introduces a voxel branch using sparse convolutions and CNNs to reduce cross-modal discrepancies and enhance matching accuracy. However, voxelization and downsampling still result in loss of original spatial information. 

To address this problem, we retain crucial spatial information by extracting edge points and pixels from original point clouds and images, preserving the complete spatial details of edge points and pixels to significantly enhance registration accuracy. To this end, we combine points showing abrupt changes in depth (depth discontinuities) with those showing sudden variations in reflectance (reflectance discontinuities). This approach improves robustness across different scene types. For edge pixels in images, we use the line segment detector (LSD) algorithm \cite{4731268} to maintain similarity with point cloud edge points while managing the number of extracted pixels. 

Validation using the KITTI dataset demonstrates the superior accuracy and robustness of our method, effectively preserving original data and improving the accuracy of cross-modal registration.
\begin{figure}[H]
    \centering
    \includegraphics[width=1\linewidth]{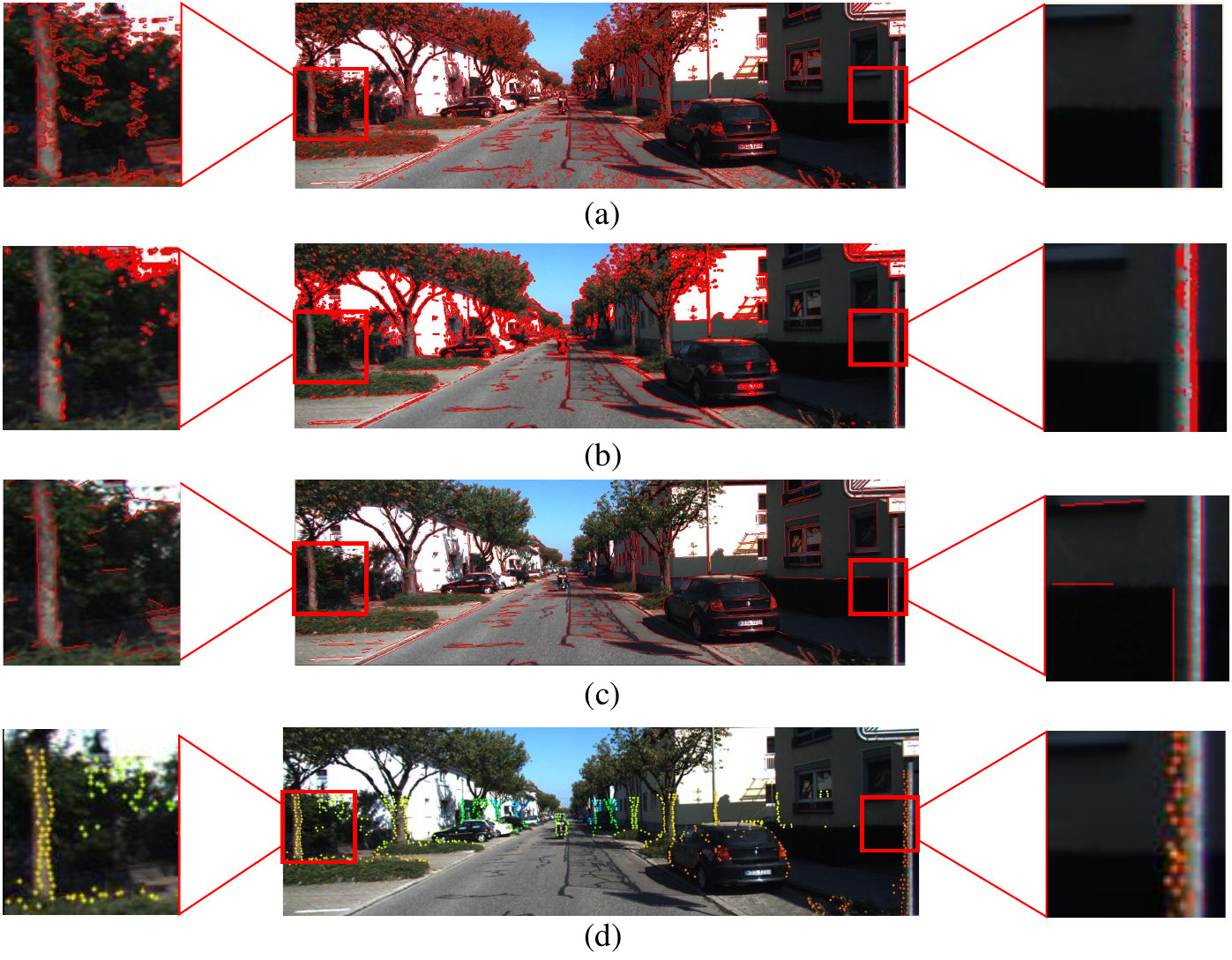}
    \caption{Visual comparison of various edge extraction methods on the KITTI Odometry dataset. This is juxtaposed with the visualization of 3D edge point projection images. (a) Edge extraction using the Canny operator with threshold values of (50, 150) yields 63406 edge pixels. (b) Edge extraction using the Sobel operator with threshold values of (0, 150) results in 61176 edge pixels. (c) Edge extraction using the LSD algorithm produces 20933 edge pixels. (d) Visualization of depth-discontinuous points obtained by projecting pre-processed point clouds onto the image plane.}
    \label{f4}
\end{figure}
\subsubsection{2D Edge Pixels Extraction\label{4A1}}
To ensure sufficient overlap between the key pixels extracted from the 2D images and the 3D edge points, while also avoiding a significant increase in computational load, we extract edges from the images and use the resulting edge pixels as 2D key pixels. Many methods are available, including gradient-based techniques such as the Sobel operator \cite{5233425} and the Canny edge detection algorithm \cite{4767851}. Specifically, we use the LSD algorithm \cite{4731268} to extract line segments from the grayscale image. All pixels falling on the extracted lines are considered as edge pixels in the image, denoted as \(kp_{2D}\in\mathbb{N}^{N_{2D}\times 2}\). As illustrated in Figure \ref{f4}, the LSD algorithm gives fewer and more significant points than the Sobel operator and the Canny edge detector \cite{pautrat2023gluestick}.
\begin{figure*}[htbp] 
    \centering
    \includegraphics[width=0.9\linewidth]{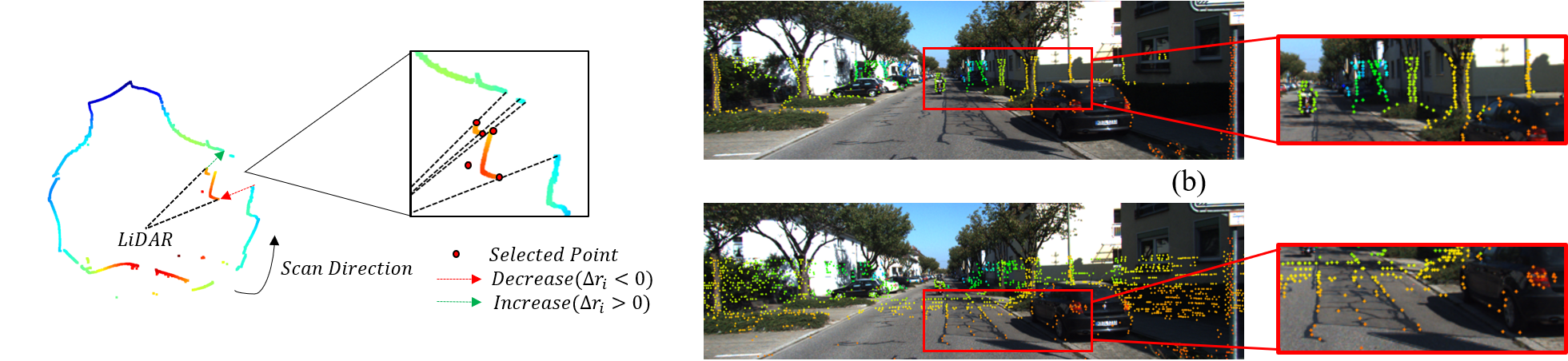}
    \caption{Demonstration of the selection of depth-discontinuous points and visualization of their projection onto the image plane. (a) Top view of a single scan line from the point cloud collected by a 64-line LiDAR in the KITTI dataset, showing the LiDAR position and scanning direction, with depth-discontinuous points highlighted. Positions where the point radius decreases are indicated by red arrows, while positions where the radius increases are indicated by green arrows. (b) Visualization of the projection of selected depth-discontinuous points from the point cloud onto the image plane. (c) Visualization of the projection of selected reflectance-discontinuous points from the point cloud onto the image plane.}
    \label{f3}
\end{figure*}
\subsubsection{3D Edge Points Extraction\label{4A2}}
For the LiDAR point cloud, characterized by circular scanning, we use a scheme that involves depth-discontinuous points and reflectance-discontinuous points.

\textbf{Depth-discontinuous points}. 

To obtain depth-discontinuous points, the point cloud is projected onto a spherical coordinate system. The distance between the LiDAR sensor and the \(i\)-th point \((x_i, y_i, z_i)\)  is  \(r_i=\sqrt{x_i^2+y_i^2+z_i^2}\). Significant changes in distance may occur when the LiDAR circular scan reaches specific positions. In such cases, the difference between the radius of the current point and the radius of the previous point, \(\Delta r_i=r_{i+1}-r_i\), may become significantly large or small. These points are considered as depth-discontinuous points. To avoid missing depth-discontinuous points with small radii due to mechanical offsets of the LiDAR device, we use the following procedure. Let \(\epsilon_D\) be a threshold. If \(\frac{\Delta r_i}{r_i}>\epsilon_D\), the \((i-1)\)-th point will be selected as depth-discontinuous point, and if \(\frac{\Delta r_i}{r_i}<-\epsilon_D\), the \((i+1)\)-th point will be selected(Figure \ref{f3}).

\textbf{Reflectance-discontinuous points}.

Similar to depth-discontinuous points, the difference in reflectance \(R\) between the current point and the previous point is calculated for points in the circular scan, denoted as \(\Delta R_i=R_{i+1}-R_i\). We also set a threshold \(\epsilon_R\) for the difference between consecutive points. If \(\Delta R_i>\epsilon_R\), the \((i-1)\)-th point will be selected as a reflectance-discontinuous point. Conversely, if \(\Delta R_i<-\epsilon_R\), the \((i+1)\)-th point will be selected.

Finally, we take the union of the sets of depth-discontinuous points and reflectance-discontinuous points as the set of 3D edge points \(kp_{3D}\in\mathbb{R}^{N_{3D}\times 4}\). Particularly noteworthy is that, in order to ensure the ratio of actual matching points, for \(kp_{2D}\), since LSD  generally does not require threshold adjustment, we directly process the image using the default threshold. For the depth discontinuities in \(kp_{3D}\), we typically set the decision threshold \(\epsilon_D\) at 0.1, meaning that a change exceeding one-tenth of the distance value is considered a depth discontinuity. Similarly, for reflectance, since it is already a normalized value, we set \(\epsilon_R\) at 0.2. Under this setting, the ratio of actual matching points to the total can exceed 10\%, which is more conducive to training.

\subsection{Feature Extraction\label{4B}}

The choice of feature extractor is influenced by the inherent modal differences between 2D and 3D data. Matching these data modalities typically involves leveraging the similarity of high-dimensional features. Initially, this entails using neural network-based feature extractors. In the feature extractor component, the prevailing methods like DeepI2P and CorrI2P partition the feature extraction process into two branches: the image branch and a point cloud brach.

The image branch typically uses CNNs to extract features from images, obtaining feature maps at smaller scales. On the other hand, the point cloud branch uses downsampling to obtain high-dimensional features from sparser point clouds \cite{an2023sp}. However, downsampling results in the loss of original position for both the point cloud and the image data. Attempting to restore high-dimensional features corresponding to the original inputs via upsampling would consume substantial memory resources and increase computational complexity.
\begin{figure*}[htbp] 
    \centering
    \includegraphics[width=0.97\linewidth]{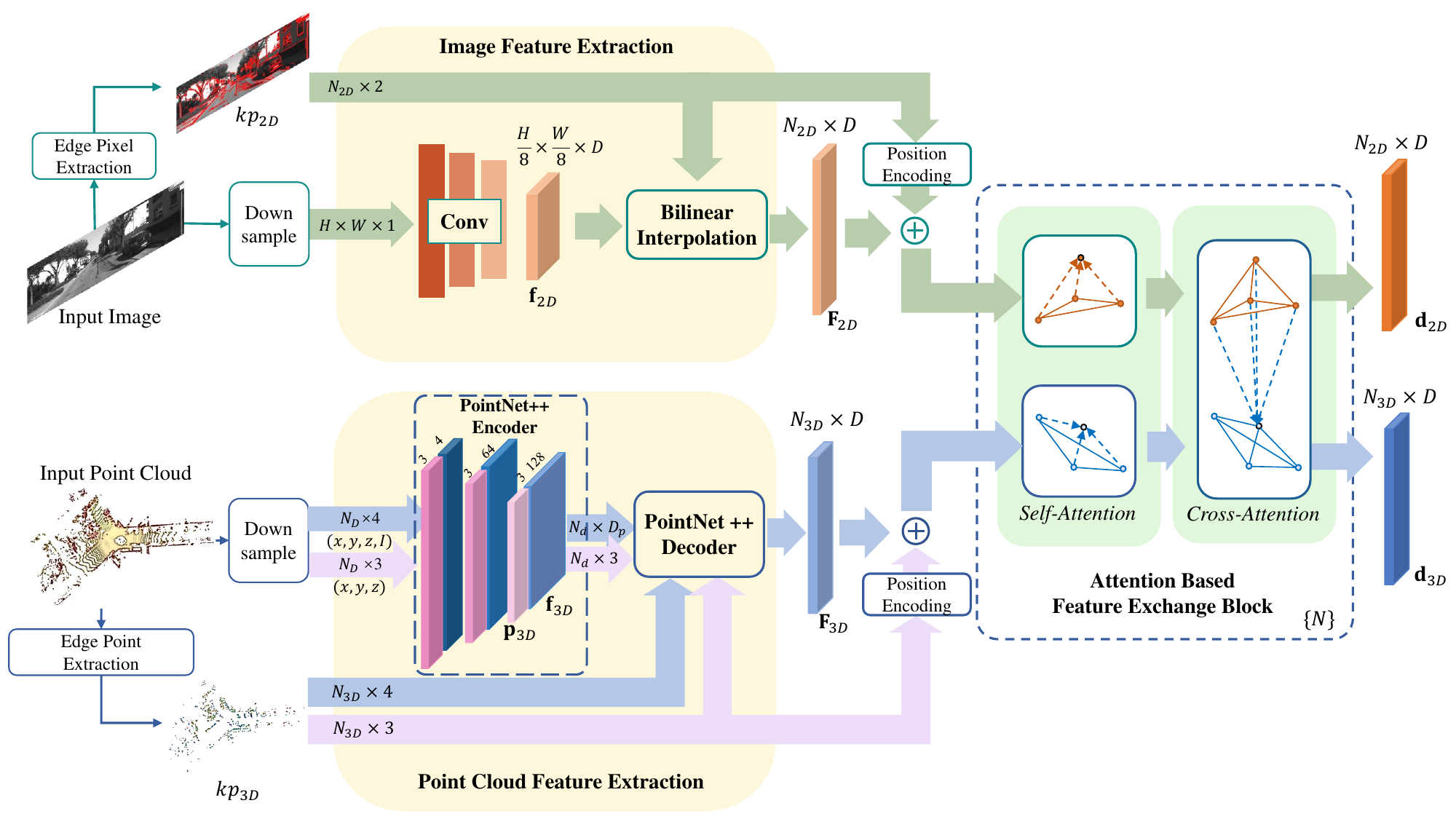}
    \caption{EdgeRegNet Structure. EdgeRegNet consists of two main components: the feature extraction module on the left and the Attention-Based Feature Exchange Block on the right. The feature extraction module includes image feature extraction and point cloud feature extraction. These regions extract high-dimensional features \(\mathbf{F}_{2D}\) and \(\mathbf{F}_{3D}\) from the preprocessed data, representing 2D and 3D feature points, respectively. The extracted features are then fed into the Attention-Based Feature Exchange Block after position embedding, producing updated features \(\mathbf{d}_{2D}\) and \(\mathbf{d}_{3D}\). The {N} in the diagram denotes the consecutive use of N Attention-Based Feature Exchange Blocks.}
    \label{f5}
\end{figure*}
To address this challenge, we propose a new approach. First, we extract global features from the entire image and point cloud data. Then, we aggregate the extracted global features onto edge feature points using specific methods. This approach preserves the original spatial information, while reducing overall memory and computational complexity.\\

\subsubsection{Point Cloud Feature Extraction}
We use PointNet++ as the point cloud feature extractor(Fig. \ref{f5}, bottom-left). It uses two layers of feature encoders to capture global features from the input point cloud. Following downsampling via farthest point sampling (FPS) within the network, the PointNet++ encoder yields the downsampled point cloud \(\mathbf{p}_{3D}\in\mathbb{R}^{N_d\times{3}}\) along with its corresponding global feature \(\mathbf{f}_{3D}\in\mathbb{R}^{N_d\times{D_p}}\). Subsequently, these features are aggregated onto nearby edge points based on Euclidean distance. PointNet++ uses its feature propagation module during upsampling to aggregate the features of the edge points, which include coordinate information and reflectance, with those of the points in the proximity of \(\mathbf{p}_{3D}\), resulting in the features on the edge points \(\mathbf{F}_{3D} \in \mathbb{R}^{N_{3D} \times D}\). That is, we use the method from the PointNet++ decoder to obtain features at specific points. By replacing the dense point cloud positions and features, which the decoder's upsampling process originally required, with the positions and features of the extracted edge points, we can conveniently aggregate global feature information to the edge points.
\subsubsection{Image Feature Extraction}
For image extraction, we use a CNN architecture similar to SuperPoint \cite{detone2018superpoint}(Fig. \ref{f5}, top-left). Upon traversing through convolutional layers, we acquire a high-dimensional image feature \(\mathbf{f}_{2D}\in\mathbb{R}^{\frac{W}{8}\times\frac{H}{8}\times{D}}\). We do not use a neural network to select feature points from the image. Instead, we directly designate edge pixels, extracted using the LSD algorithm, and apply bilinear interpolation to aggregate global feature information to the edge points. This requires the position information of edge pixels as additional input. This facilitates the acquisition of features \(\mathbf{F}_{2D}\in\mathbb{R}^{N_{2D}\times{D}}\) corresponding to the edge pixels within the 2D image.
\subsection{Attention-Based Feature Exchange Block\label{4C}}
After feature extraction, we obtain the features \(\mathbf{F}_{2D}\) and \(\mathbf{F}_{3D}\) for 2D edge pixels and 3D edge points, respectively. However, because of their different feature extraction methods and the existence of cross-dimensional differences, a direct comparison of high-dimensional features between the two is not feasible. The VP2P-Match paper \cite{zhou2023differentiable} also notes this discrepancy. It attributes it to the fact that the features obtained by PointNet++ are derived from MLP networks, whereas the 2D features are obtained through 2D convolutional neural networks. This results in a significant mismatch in feature space. Unlike methods such as DeepI2P \cite{li2021deepi2p} and CorrI2P \cite{Ren_2023}, which fuse multichannel features from an image grid and a downsampled point cloud, we propose using a point-based feature exchange module. This module reduces cross-modal discrepancies by exchanging features between image edge pixels and point cloud edge points. It uses an attention-based feature exchange block similar to the GNN module in SuperGlue \cite{sarlin2020superglue} to reduce the impact of these differences. This is achieved by enabling information exchange and updating through self-attention and cross-attention mechanisms between features originating from different dimensions and extracted by different feature extractors.

First, we encode the positions of the input 2D and 3D features(Fig. \ref{f5}, right half). After positional encoding, we input cross-dimensional features into the attention-based feature exchange block. In this block, we aggregate and exchange feature information from cross-modal features \(\mathbf{F}_{2D}\) and \(\mathbf{F}_{3D}\) through the attention mechanism. The self-edges are based on self-attention, and the cross-edges are based on cross-attention.
\begin{figure*}[htbp] 
    \centering
    \includegraphics[width=0.65\linewidth]{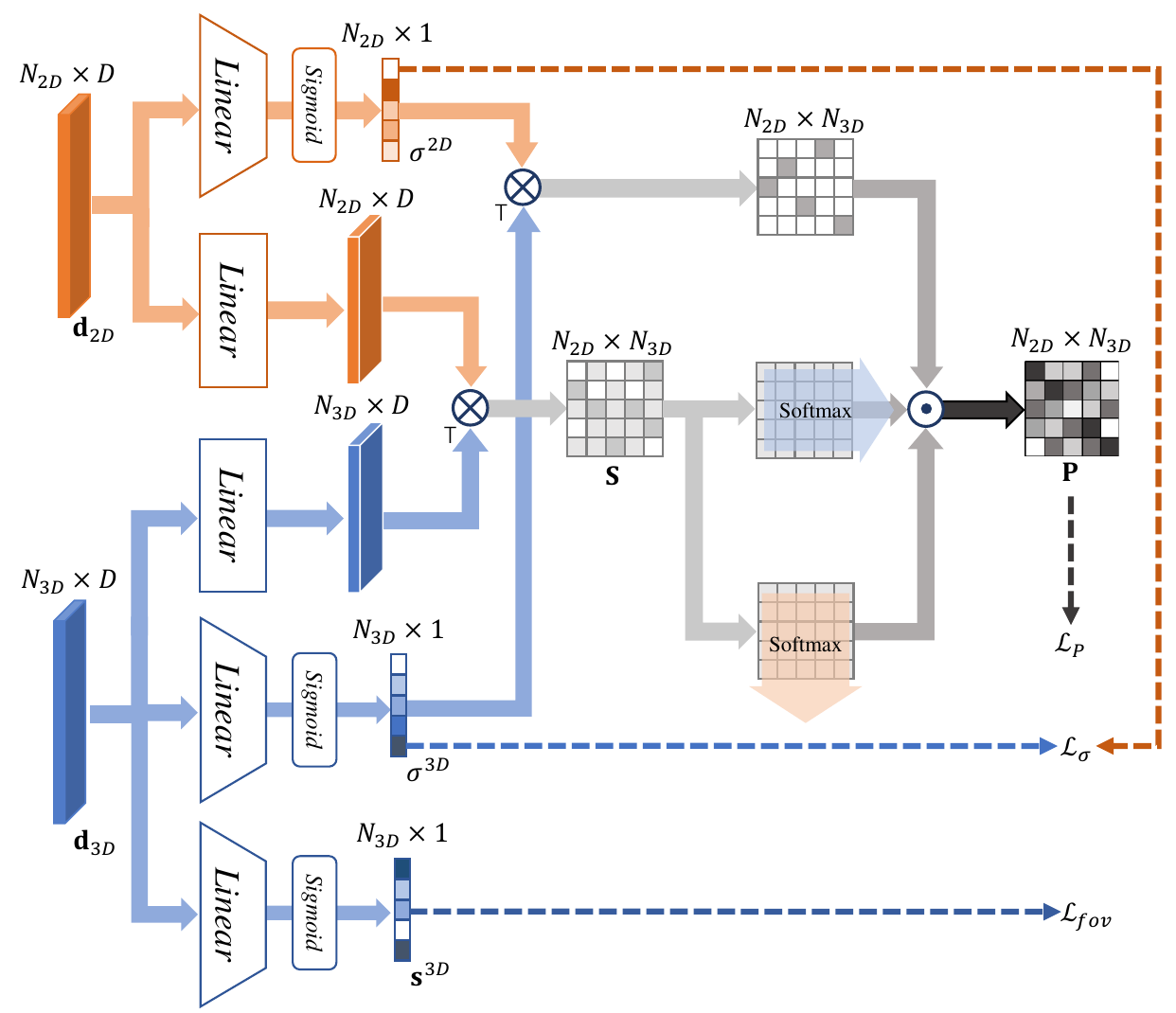}
    \caption{Optimal Matching Layer, showcasing the workflow from the edge features \( \mathbf{d}_{2D} \) and \( \mathbf{d}_{3D} \) to the correspondence determination using the partial assignment matrix \( \mathbf{P} \). The linear blocks denote neural network layers with different output sizes, and the 1D Conv blocks represent one-dimensional convolution operations. Softmax operations applied across matrix rows and columns are represented by arrows in distinct colors. Additionally, the positions where various loss functions are applied are annotated in the diagram.}
    \label{f6}
\end{figure*}
We use the input features \(\mathbf{F}_{2D}\) and \(\mathbf{F}_{3D}\)to calculate the query (Q), key (K), and value (V) matrices, which are usually obtained by linear transformations. Taking \(\mathbf{F}_{2D}\) as an example, we can calculate its corresponding three matrices as follows.
\[
\begin{cases}
\mathbf{Q}_m = \mathbf{F}_m \mathbf{W}_Q, \\
\mathbf{K}_m = \mathbf{F}_m \mathbf{W}_K, \\\tag{4}
\mathbf{V}_m = \mathbf{F}_m \mathbf{W}_V,
\end{cases}
\]where \(\mathbf{W}_Q\), \(\mathbf{W}_K\), and \(\mathbf{W}_V\) are the weight matrices for the query, key, and value, respectively. Using the same method, we can calculate the corresponding \(\mathbf{Q}_{3D}\), \(\mathbf{K}_{3D}\), and \(\mathbf{V}_{3D}\) for \(\mathbf{F}_{3D}\).\\
\textbf{Self-attention.} In the self-attention mechanism, each element within a set of characteristics attends to all other elements within the same set. The attention scores for \(\mathbf{F}_{2D}\) are calculated by taking the dot product of the query and key matrices, followed by a softmax function:
\[
\text{Attention}_m = \text{softmax}\left(\frac{\mathbf{Q}_m \mathbf{K}_m^\mathsf{T}}{\sqrt{d_k}}\right)\mathbf{V}_m,\tag{5}
\]where \(d_k\) is the dimensionality of the key vectors.\\
\textbf{Cross-attention.} The cross-attention scores are calculated similarly to self-attention but between different sets of features.\[
\text{Cross-Attention}_{m \rightarrow n} = \text{softmax}\left(\frac{\mathbf{Q}_m \mathbf{K}_n^\mathsf{T}}{\sqrt{d_k}}\right)\mathbf{V}_n,\tag{6}\]where \(m\) amd \(n\) are cross-modal pairs. The resulting self-attention feature representations are further processed by the cross-attention module. This mechanism facilitates the exchange of information between two feature sets, capturing their correlations and updating each other's information. The inputs and outputs of the attention-based feature exchange network have identical dimensions, producing updated 2D and 3D features \(\mathbf{d}_{2D}\in\mathbb{R}^{N_{2D}\times{D}}\) and \(\mathbf{d}_{3D}\in\mathbb{R}^{N_{3D}\times{D}}\).

To enhance the feature expression capabilities and performance, six stacked attention-based feature exchange network modules are cascaded. After passing through the attention-based feature exchange network, we obtain the final features  \(\mathbf{d}_{2D}\) and\(\mathbf{d}_{3D}\), from the entire feature extraction stage. These two features will be used in the subsequent matching process. 
\subsection{Optimal Matching Layer\label{4D}}
Mainstream methods usually rely on comparing cosine similarities of high-dimensional features to establish correspondences between 2D and 3D points. In contrast, our approach, aiming to overcome inherent cross-modal differences and enhance the quantity and accuracy of matched point pairs, integrates the optimal matching layer from LightGlue \cite{lindenberger2023lightglue}. 
This deep feature matching method uses the optimal transport problem to solve a partial assignment problem, effectively enhancing the robustness of the matching algorithm. Initially introduced by SuperGlue in the registration domain, this method uses the Sinkhorn algorithm for computation, involving multiple iterations of normalization along rows and columns, which reduces computational efficiency. LightGlue has improved upon this by separating similarity and matchability computations, thus enhancing computational efficiency while ensuring cleaner gradients in the computation process. 

As shown in Figure \ref{f6}, we introduce a matching optimization layer that is more suitable for our task. First, we compute pairwise scores between the 2D edge pixels \(kp_{2D}\) and the 3D edge points \(kp_{3D}\) indexed by using a learned linear transformation with bias. The linear layer here has the same number of input and output channels:
\[
\mathbf{S} = \text{Linear}\left(\mathbf{d}_{2D}\right)\text{Linear}\left(\mathbf{d}_{3D}\right) ^\mathsf{T}.\tag{7}
\]
The elements in matrix \(\mathbf{S}\in\mathbb{R}^{N_{2D}\times N_{3D}}\) represent the similarity between each 2D edge pixel and each 3D edge point. The linear layer here has the same number of input and output channels. Specifically, we have \(\text{Linear}\left(\mathbf{d}_{2D}\right)\in \mathbb{R}^{N_{2D}\times D}\) and \(\text{Linear}\left(\mathbf{d}_{3D}\right)\in\mathbb{R}^{N_{3D}\times{D}}\). Additionally, we separately calculate the matching scores for each individual 2D pixel and 3D point:
\[
\mathbf{\sigma}^{2D} = \text{Sigmoid}\left(\text{Linear}\left(\mathbf{d}_{2D}\right)\right) ,\tag{8}
\]
\[
\mathbf{\sigma}^{3D} = \text{Sigmoid}\left(\text{Linear}\left(\mathbf{d}_{3D}\right)\right) .\tag{9}
\]
The linear layer here has a single output channel. That is, we have \(\text{Linear}\left(\mathbf{d}_{2D}\right)\in \mathbb{R}^{N_{2D}\times 1}\) and \(\text{Linear}\left(\mathbf{d}_{3D}\right)\in\mathbb{R}^{N_{3D}\times{1}}\). These scores indicate the probability of each pixel or point having a corresponding match. 

Next, we integrate both similarity and matchability scores into a soft partial assignment matrix \(\mathbf{P}\), where:
\[
\mathbf{P}_{ij} = \mathbf{\sigma}^{2D}_{i} \cdot \mathbf{\sigma}^{3D}_{j} \cdot \frac{\exp(\mathbf{S}_{ij})}{\sum^{N_{2D}}_{k =1} \exp(\mathbf{S}_{kj})} \cdot \frac{\exp(\mathbf{S}_{ij})}{\sum^{N_{3D}}_{k=1} \exp(\mathbf{S}_{ik})}.\tag{10}
\]
A correspondence between the \(i\)-th pixel and the \(j\)-th point is established when both are predicted to be matchable and their similarity exceeds that of any other pair in matrix \(\mathbf{P}\). Specifically, we select pairs \((i,j)\) such that \(\mathbf{P}_{ij}\) is the largest element in both row \(i\) and column \(j\) of matrix \(\mathbf{P}\).

After obtaining the 2D-3D matching point pairs, the third step involves using the EPnP algorithm with RANSAC to estimate the transformation matrix of the given 2D-3D point pairs. The EPnP algorithm takes a set of known 2D image points and corresponding 3D space points as input and outputs the transformation matrix \(\mathbf{T}\)(see (1)).
\subsection{Loss Function}
To train the proposed network, we use a loss function consisting of three components: \(L_{fov}\), \(L_\mathbf{\sigma}\), and \(L_P\).

During the computation of the loss function, 3D edge points are projected onto the camera plane using the transformation matrix from the ground truth and the intrinsic parameters of the camera, yielding the 2D projections of the 3D edge points. We set a threshold \(\epsilon_c\) to establish the correspondence between the 3D edge points and the 2D edge pixels. If the Euclidean distance between the 2D projected points and a 2D edge pixel is less than \(\epsilon_c\), then we consider the 3D edge point corresponding to the 2D projected point and the 2D edge pixel as a ground-truth matching pair. Based on these actual pairs of matching points, we design the following loss functions.

First, \(L_{fov}\) is computed. The field of view (FOV) of the LiDAR-scanned point cloud generally covers the entire scene, while the camera typically captures only a small portion of the entire scene. Filtering the point cloud by predicting whether its points are within the camera's FOV can reduce mismatches. The calculation starts by passing the feature of 3D feature points through a 1D convolutional layer, followed by a sigmoid activation function, which outputs a predicted score \(\mathbf{s}^{3D}\). Ground truth information \(\hat{\mathbf{s}}^{3D}\), which indicates whether 3D edge points will be projected on the image plane, is recorded when projecting the 3D feature points onto the camera plane. The binary cross-entropy between these two scores is then computed to obtain \(L_{fov}\):
\[L_{fov} = - \frac{1}{N_{3D}} \sum_{i=1}^{N_{3D}}[\hat{\mathbf{s}}^{3D}_i \cdot \log{\mathbf{s}^{3D}_i} + (1 - \hat{\mathbf{s}}^{3D}_i) \cdot \log{(1 - \mathbf{s}^{3D}_i)}].\tag{11}\]
After projecting the 3D point cloud onto the image plane, a radius threshold \(\epsilon_c\) is set. If the Euclidean distance between the projected point of a 3D feature point and a certain 2D feature point is less than \(\epsilon_c\), then the two points are considered as 2D-3D matching point pairs. After obtaining the pairs of matching points, the situations of whether the 2D and 3D points are matching points are recorded as \(\mathbf{\hat{\sigma}}^{2D}\) and \(\mathbf{\hat{\sigma}}^{3D}\), respectively. \(\mathbf{\hat{\sigma}}^{3D}\) and \(\mathbf{\hat{\sigma}}^{2D}\) are then used to compute the matching binary cross-entropy loss functions \(L_{\mathbf{\sigma}1}\) and \(L_{\mathbf{\sigma}2}\):
\[L_{\mathbf{\sigma}1} = - \frac{1}{N_{2D}} \sum_{i=1}^{N_{2D}} [\mathbf{\hat{\sigma}}^{2D}_i \cdot \log{\mathbf{\sigma}^{2D}_i} + (1 - \mathbf{\hat{\sigma}}^{2D}_i) \cdot \log{(1 - \mathbf{\sigma}^{2D}_i)}], \tag{12}\]
\[L_{\mathbf{\sigma}2} = - \frac{1}{N_{3D}} \sum_{i=1}^{N_{3D}}[\mathbf{\hat{\sigma}}^{3D}_i \cdot \log{\mathbf{\sigma}^{3D}_i} + (1 - \mathbf{\hat{\sigma}}^{3D}_i) \cdot \log{(1 - \mathbf{\sigma}^{3D}_i)}]. \tag{13}\]
The sum of \( L_{\mathbf{\sigma}1}\) and \(L_{\mathbf{\sigma}2}\) is denoted by \(L_\mathbf{\sigma} \).

Regarding the partial assignment matrix \(\mathbf{P}\), the positions corresponding to the maximum values in the rows and columns are considered as a predicted 2D-3D matching pair. After obtaining the actual pairs, for the actual matching positions, the values corresponding to the positions in the partial assignment matrix \(\mathbf{P}\) should be relatively large. Based on this fact, \(L_P\) is calculated as:
\[L_P = -\frac{1}{|\hat{\mathcal{M}}|} \sum_{(i,j)\in \hat{\mathcal{M}}} \log{\mathbf{P}_{ij}},\tag{14}\]where \(\hat{\mathcal{M}}\) represents the set of positions of the actual pairs of matching points.

The overall loss function is the weighted sum of these three components:

\[L = \lambda_{fov} L_{fov} + \lambda_\mathbf{\sigma} L_\mathbf{\sigma} + \lambda_P L_P. \tag{15}\]

\section{Experimental Results and Analysis }
\begin{table*}[ht]

\centering
\caption{Comparison of Registration Accuracy across Different Methods on KITTI Odometry and nuScenes Datasets. ``↓'' indicates that lower values are better, and ``↑'' indicates that higher values are better. The best results are highlighted in bold.}
\label{t1}
\resizebox{\textwidth}{!}{
\begin{tabular}{ll|ccllc|cclcl}
\hline
  &&
   \multicolumn{5}{c|}{KITTI}&
   \multicolumn{5}{c}{nuScenes}\\  \cline{3-12}Method&&
  \multicolumn{1}{c|}{RTE(m) ↓} &
  \multicolumn{1}{c|}{RRE(°) ↓} &
    &Acc.↑ 
&&
  \multicolumn{1}{c|}{RTE(m) ↓} &
  \multicolumn{1}{c|}{RRE(°) ↓} &
   &Acc.↑  &\\ \hline
 Grid Cls. + PnP \cite{li2021deepi2p}
&&
  \multicolumn{1}{c|}{1.07 ± 0.61} &
  \multicolumn{1}{c|}{6.48 ± 1.66} &
    &11.22 
&&
  \multicolumn{1}{c|}{2.35 ± 1.12} &
  \multicolumn{1}{c|}{7.20 ± 1.65} &
   &2.45  &\\
 DeepI2P (3D) \cite{li2021deepi2p}
&&
  \multicolumn{1}{c|}{1.27 ± 0.80} &
  \multicolumn{1}{c|}{6.26 ± 2.29} &
    &3.77 
&&
  \multicolumn{1}{c|}{2.00 ± 1.08} &
  \multicolumn{1}{c|}{7.18 ± 1.92} &
   &2.26  &\\
 DeepI2P (2D) \cite{li2021deepi2p}
&&
  \multicolumn{1}{c|}{1.46 ± 0.96} &
  \multicolumn{1}{c|}{4.27 ± 2.74} &
    &25.95 
&&
  \multicolumn{1}{c|}{2.19 ± 1.16} &
  \multicolumn{1}{c|}{3.54 ± 2.51} &
   &38.10  &\\
 CorrI2P \cite{Ren_2023}
&&
  \multicolumn{1}{c|}{0.74 ± 0.65} &
  \multicolumn{1}{c|}{2.07 ± 1.64} &
    &72.42 
&&
  \multicolumn{1}{c|}{1.83 ± 1.06} &
  \multicolumn{1}{c|}{2.65 ± 1.93} &
   &49.00  &\\
 VP2P-Match \cite{zhou2023differentiable}
&&
  \multicolumn{1}{c|}{0.59 ± 0.56} &
  \multicolumn{1}{c|}{2.39 ± 2.08} &
    &83.04 
&&
  \multicolumn{1}{c|}{0.89 ± 1.44} &
  \multicolumn{1}{c|}{2.15 ± 7.03} &
   &88.33  &\\ \hline
 Ours &&
  \multicolumn{1}{c|}{\textbf{0.54 ± 0.49}} &
  \multicolumn{1}{c|}{\textbf{1.65 ± 1.62}} &
    &\textbf{90.24}&&
  \multicolumn{1}{c|}{\textbf{0.68 ± 0.60}}&
  \multicolumn{1}{c|}{\textbf{1.26 ± 0.85}} &
   &\textbf{94.62}&\\ \hline
\multicolumn{12}{l}{\footnotesize  Note: VP2P-Match results on nuScenes are from their paper, as the code is not public.}\\
\end{tabular}}
\end{table*}

\subsection{Dataset}
As in \cite{Ren_2023}, we conducted experiments on the KITTI Odometry and nuScenes datasets. 

\textbf{KITTI Odometry\cite{6248074}.} We generated image-point cloud pairs from the same data frame of 2D/3D sensors. Following previous works, we used sequences 00-08 for training and sequences 09-10 for testing. For the point clouds, we applied random translations within the ±10 range on the x and y axes and unrestricted rotations around the z-axis. Unlike the DeepI2P method, our input image size was closer to the original data at 1250×376, and we extracted edge pixels directly from the input image without any additional cropping. For point cloud data, we extracted edge points from the original point clouds, which were downsampled to 20,480 points before being input to the feature extraction network.

\textbf{nuScenes\cite{caesar2020nuscenes}.} For nuScenes, we did not filter the scenes to better reflect real-world usage. We generated image-point cloud pairs from the same frame of the original 2D and 3D sensor data. We trained the network using 850 training scenes and tested it on 150 scenes from the dataset. The input images were 1600×900 pixels in size, and we directly extracted edge pixels from these images. For point clouds, we extracted edge points from the original point cloud, which was downsampled to 20,480 points before being input to the feature extraction network.
\begin{table*}[ht]
\centering
\caption{Comparison of performance with other methods. Smaller values for Model Size and Pose Inference are better. The best results are highlighted in bold.}
\label{t2}
\resizebox{\textwidth}{!}{
\fontsize{7}{8}\selectfont 
\begin{tabular}{ccc|c|c|ccc|c|ccc}
\hline
 &&& DeepI2P(2D) & DeepI2P(3D) &  & CorrI2P && VP2P-Match &  &Ours   &\\ \hline
  &Model size(MB)    
&& 100.12      & 100.12      &  & 141.07  
&& 30.73&  &\textbf{9.77}&\\
  &Inference Time(s)&& 23.47       & 35.61       &  & 8.96    && 0.19       &  &\textbf{0.18} &\\ \hline
\end{tabular}}
\end{table*}
\begin{figure*}[htbp] 
    \centering
    \includegraphics[width=0.95\linewidth]{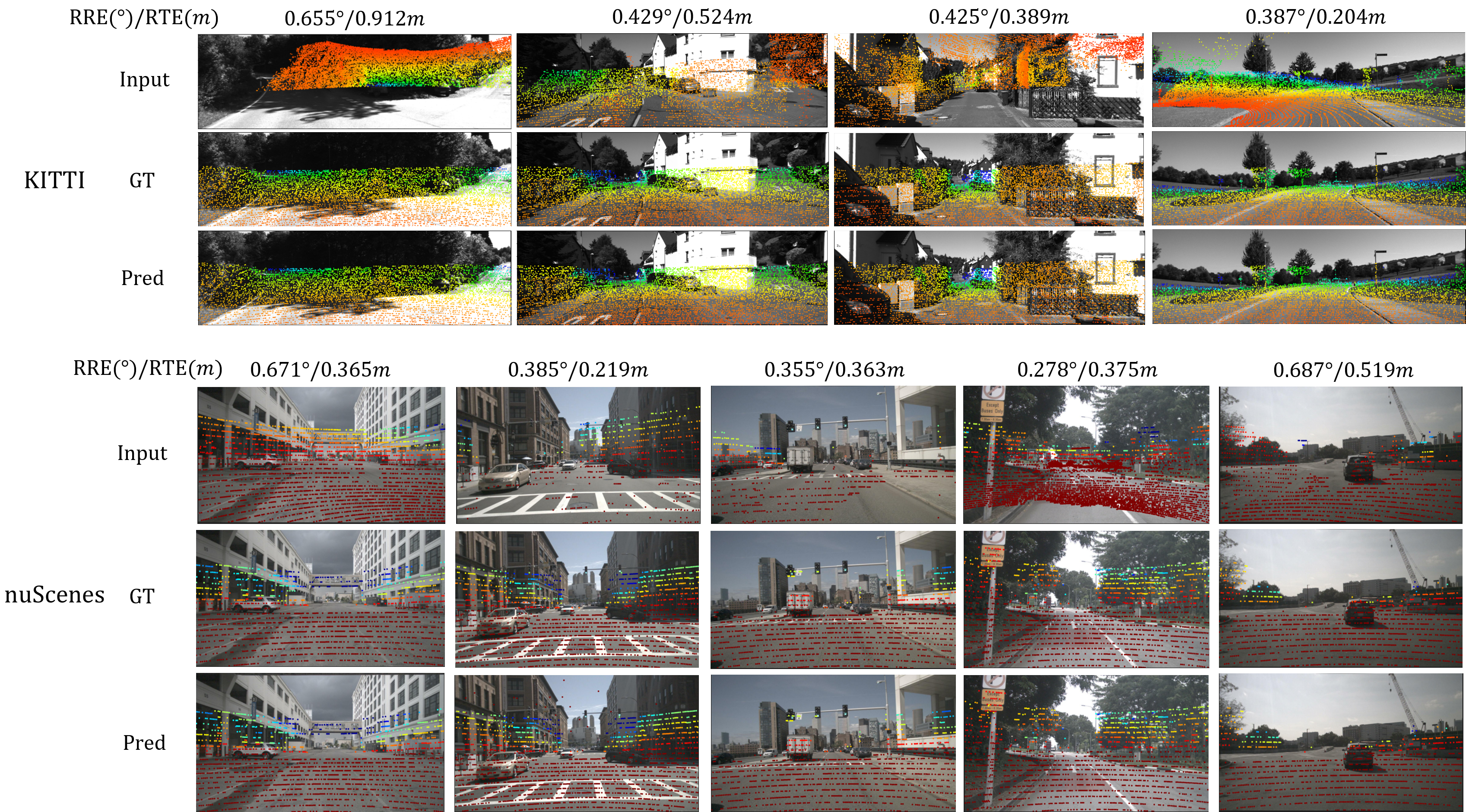}
    \caption{Visualization of registration results on the KITTI and nuScenes datasets. The top row shows results from the KITTI dataset, while the bottom row shows results from the nuScenes dataset. The RRE and RTE data for each frame are annotated above the corresponding images. The colors of the points in the images represent the distance from the points to the origin of the LiDAR coordinate system. Among them, the farther points are represented in cool colors, while the closer points are represented in warm colors.}

    \label{f7}
\end{figure*}

\subsection{Compared Methods}
We benchmarked the proposed method against three methods: DeepI2P \cite{li2021deepi2p}, CorrI2P \cite{Ren_2023}, and VP2P-Match \cite{NEURIPS2023_a0a53fef}. Notably, we only compared methods that only estimate once; some coarse-to-fine methods are not in consideration as they will make extra optimization after the first estimation.
\begin{enumerate}
    \item \textbf{DeepI2P.} This approach offers two distinct methods: \textbf{Grid Cls. + PnP} and \textbf{Frus. Cls. + Inv. Proj.}. The \textbf{Grid Cls. + PnP} method first segments the input image into a 32×32 grid. Then, it trains a neural network to classify 3D points into specific 2D grid cells. Finally, it uses the EPnP algorithm along with RANSAC to estimate the transformation matrix \(\mathbf{T}\). The \textbf{Frus. Cls. + Inv. Proj.} method introduces frustum classification using inverse camera projection to determine the transformation matrix, exploring both 2D and 3D inverse projections, referred to as DeepI2P(2D) and DeepI2P(3D), respectively.
    \item \textbf{CorrI2P.} Building upon DeepI2P, CorrI2P enhances registration accuracy by using a neural network equipped with a feature exchange module and supervision of overlapping regions. This method learns correspondences between image and point cloud pairs. The EPnP algorithm combined with RANSAC is used to predict the transformation matrix, thereby improving registration performance.
    \item \textbf{VP2P-Match.} VP2P-Match focuses on improving the accuracy and speed of pixel-to-point matching by leveraging sparse convolution to enhance the similarity between point cloud features and CNN-extracted image features. It incorporates a differentiable PnP solver into an end-to-end training framework, which facilitates the learning of a structured cross-modal latent space using adaptive weighted optimization for image-to-point cloud registration.
\end{enumerate}
\subsection{Registration Accuracy}
To evaluate registration accuracy, we used the same statistical methods as in \cite{Ren_2023} and \cite{li2021deepi2p}. Similarly to point-to-point (P2P) registration, we used the relative translational error (RTE) \(E_t\) and the relative rotational error (RRE) \(E_R\) to evaluate our registration results. These errors are computed as
\[E_R = \sum_{i=1}^{3} \left| \gamma(i) \right|,\tag{16}\]
\[E_t = \left\| \mathbf{t}_{gt} - \mathbf{t}_E \right\|,\tag{17}\]where \(\gamma(i), i=1,2,3\) are the Euler angles of the matrix \(\mathbf{R}_{gt}^{-1}\mathbf{R}_E\). Here, the rotation matrix \(\mathbf{R}_{gt}\) and the translation vector \(\mathbf{t}_{gt}\) denote the transformation of the ground truth, while matrix \(\mathbf{R}_E\) and vector \(\mathbf{t}_E\) define the estimated transformation. To mitigate the impact of outliers where some failed registrations result in exceptionally large RRE and RTE, we only considered results where \(\text{RRE} < 10\)° and \(\text{RTE} < 5\space m\). For the registration success rate (Acc.), we used the same criteria as in \cite{zhou2023differentiable}, where registrations with \(\text{RRE} < 5\)° and \(\text{RTE} < 2\space m\) are considered successful. The final results  are presented in Table \ref{t1}.
\begin{table}[H]
\centering
\caption{Effect of Different Edge Extraction Methods}
\label{t3}
\begin{tabular}{c|c|c|c|c}
\hline
3D Points& 2D Pixels& RTE(m) ↓    & RRE(°) ↓    & Acc.↑ \\ \hline
Random           & Random   & 0.89 ± 0.59 & 2.68 ± 1.99 & 74.03 \\
 Reflectance+Depth& Canny& 2.63 ± 4.61& 5.09 ± 2.53&16.51\\
 Reflectance+Depth& Sobel& 0.89 ± 0.74& 2.62 ± 1.94&76.11\\
Reflectance Only& LSD      & 0.56 ± 0.47 & 1.77 ± 1.64 & 90.15 \\
Depth Only        & LSD & 0.54 ± 0.57                           & \textbf{1.55 ± 1.65}& 88.23                           \\
Reflectance+Depth& LSD & \textbf{0.54 ± 0.49}& 1.65 ± 1.62                           & \textbf{90.24}\\ \hline
\end{tabular}
\end{table}

\begin{table}[H]
\centering
\caption{Effect of the attention-based feature exchange block }
\label{t4}
\begin{tabular}{c|c|c|c}
\hline
                  & RTE(m) ↓  & RRE(°) ↓  & Acc.↑ \\ \hline
 w/o feature exchange block& -         & -         &0     \\ 
Full& \textbf{0.54 ± 0.49}& \textbf{1.65 ± 1.62}& \textbf{90.24}\\ \hline
\multicolumn{4}{l}{\footnotesize ``-'' indicates that there is no data meeting the filtering criteria.}\\
\end{tabular}
\end{table}
Our method outperformed existing methods on both the KITTI Odometry and nuScenes datasets. Since the VP2P-Match training and testing data, as well as the code for the nuScenes dataset are not publicly available, we were unable to filter their RTE and RRE according to the standards of this paper. Therefore, we did not include a comparison on nuScenes with VP2P-Match. 

\subsection{Efficiency Comparison}
We compared the complexity of the proposed method with that of the other methods on a platform equipped with an NVIDIA GeForce RTX 3090 GPU and an Intel(R) Core(TM) i9-10900X CPU @ 3.70GHz, as shown in Table \ref{t2}. We evaluated the performance of our model using the same evaluation metrics as in \cite{zhou2023differentiable}, and compared our results based on the evaluation data provided in \cite{zhou2023differentiable}. We conducted an efficiency test on the KITTI dataset, calculating the inference time per frame by dividing the total inference time by the total number of input frames. As shown in Table \ref{t2}, our model achieved the lowest inference time while maintaining a similar model size to the best previous model.
\subsection{Visual Comparison}
Fig. \ref{f7} presents our registration results on the KITTI and nuScenes datasets, displaying the projections of the point clouds onto the image plane. The color of the points in the images represents the distance from the camera. We show the initial unaligned projection results (Input), the aligned projection results obtained using the ground truth (GT), and the projection results after registration using our predicted alignment (Pred). Our method achieved accurate registration in different environments.

\subsection{Ablation Study}
To evaluate the effectiveness of our method, we conducted two sets of ablation experiments. The first set assessed the impact of using different types of edge point in the point cloud, while the second set examined the significance of the attention-based feature exchange module.
\subsubsection{Ablation Study on Edge Extraction}
We validate the effectiveness of using edge features for registration through edge extraction ablation experiments while also evaluating the advantages and disadvantages of different edge extraction methods. Depth-discontinuous points in the point cloud emphasize spatial position information, whereas reflectance-discontinuous points emphasize optical information similar to the camera imaging process. Table \ref{t3} shows the results. ``Random'' refers to not using any edge extraction algorithm and instead using randomly selected points for matching. For 2D pixels, we analyze the performance of using the Sobel operator \cite{5233425} and the Canny edge detection algorithm \cite{4767851}. For 3D points, we analyze the scenarios of using only depth-discontinuous points, only reflectance-discontinuous points, and a combination of both.

When edge information was not used for registration, both accuracy and success rate significantly decreased. Moreover, using a single type of input source led to lower accuracy compared to using multiple input sources. This result indicates that using edge information helps improve system accuracy. Moreover, for the diverse scenes within the KITTI dataset, using multisource edge information enhanced registration robustness, improving overall registration accuracy.

\subsubsection{Ablation Study on the Attention-based Feature Exchange Module}
The attention-based feature exchange module mitigates discrepancies between cross-modal features by leveraging self-attention to refine intra-modal features and cross-attention to fuse inter-modal information. As shown in Table IV, removing this module leads to a significant performance drop, underscoring its critical role in robust cross-modal registration. While the module is highly effective, there is room for optimization in practical deployment, such as reducing computational complexity through lightweight attention mechanisms or pruning redundant feature interactions to enhance efficiency without compromising performance.

\section{Conclusion}
We explored the feasibility of using edge features for cross-modal registration and demonstrated the effectiveness of an attention-based feature exchange block in this context. Using deep learning techniques, our method effectively addresses the inherent differences between 2D images and 3D point clouds. Experiments on the KITTI and nuScenes datasets showed significant improvements in registration accuracy and computational efficiency over other methods, highlighting the robustness and effectiveness of our approach. However, our network has a large number of model parameters due to the extensive size of the feature exchange network. In the future, we will explore other more lightweight and advanced networks to further improve the performance of our model. 
\small
\bibliographystyle{IEEEbib}
\bibliography{references}
\begin{IEEEbiography}
[{\includegraphics[width=1in,height=1.25in,clip,keepaspectratio]{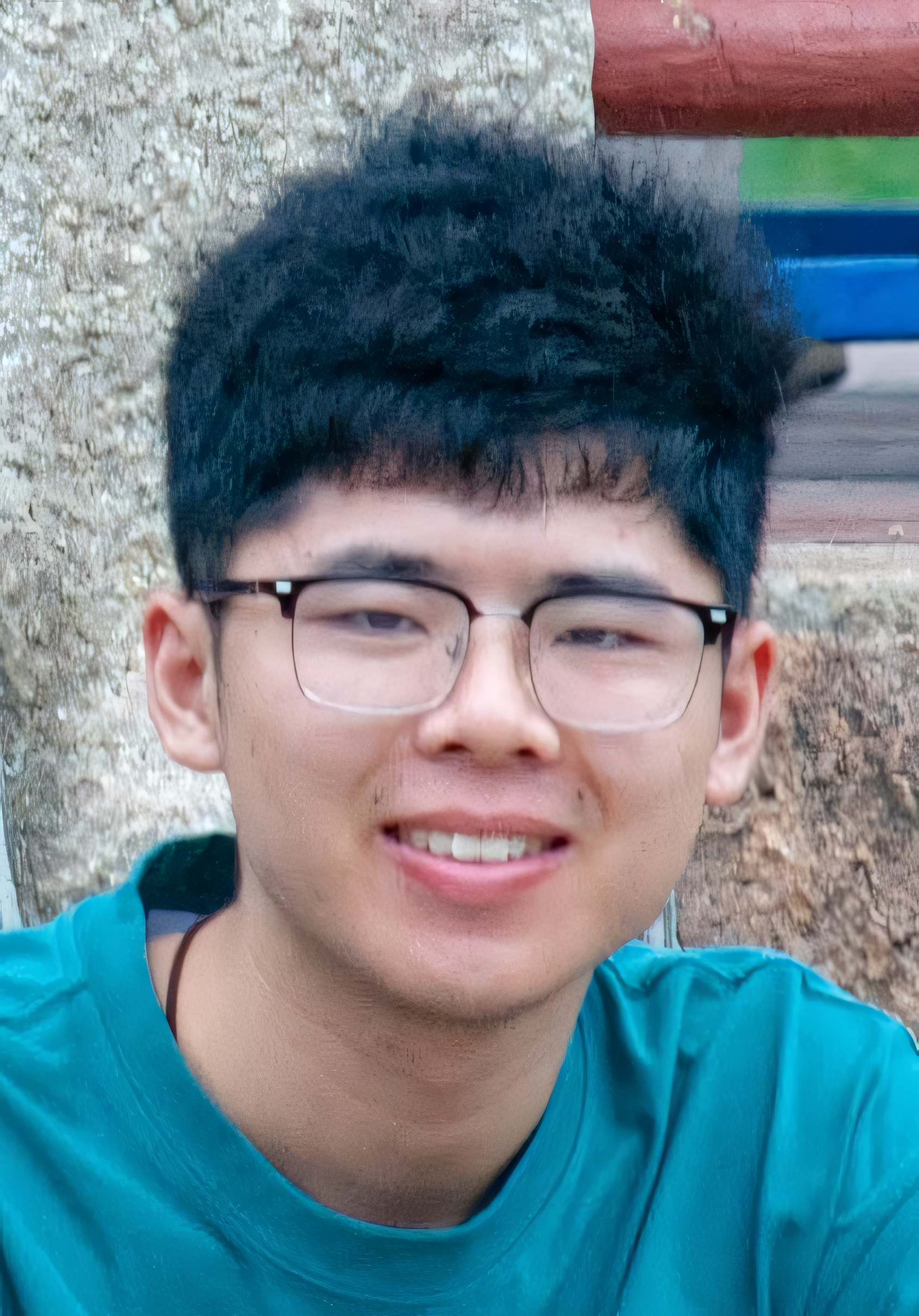}}] 
{Yuanchao Yue} received the B.S. degree in automation from Yanshan University, Hebei, China, in 2022. He is currently working toward the M.S. degree with the School of Control Science and Engineering, Shandong University, Jinan, China. His current research interests include cross-modal registration, point cloud registration, medical data processing, and multi-sensor fusion.					                 
\end{IEEEbiography}
\vspace{-1cm}
\begin{IEEEbiography}
[{\includegraphics[width=1in,height=1.25in,clip,keepaspectratio]{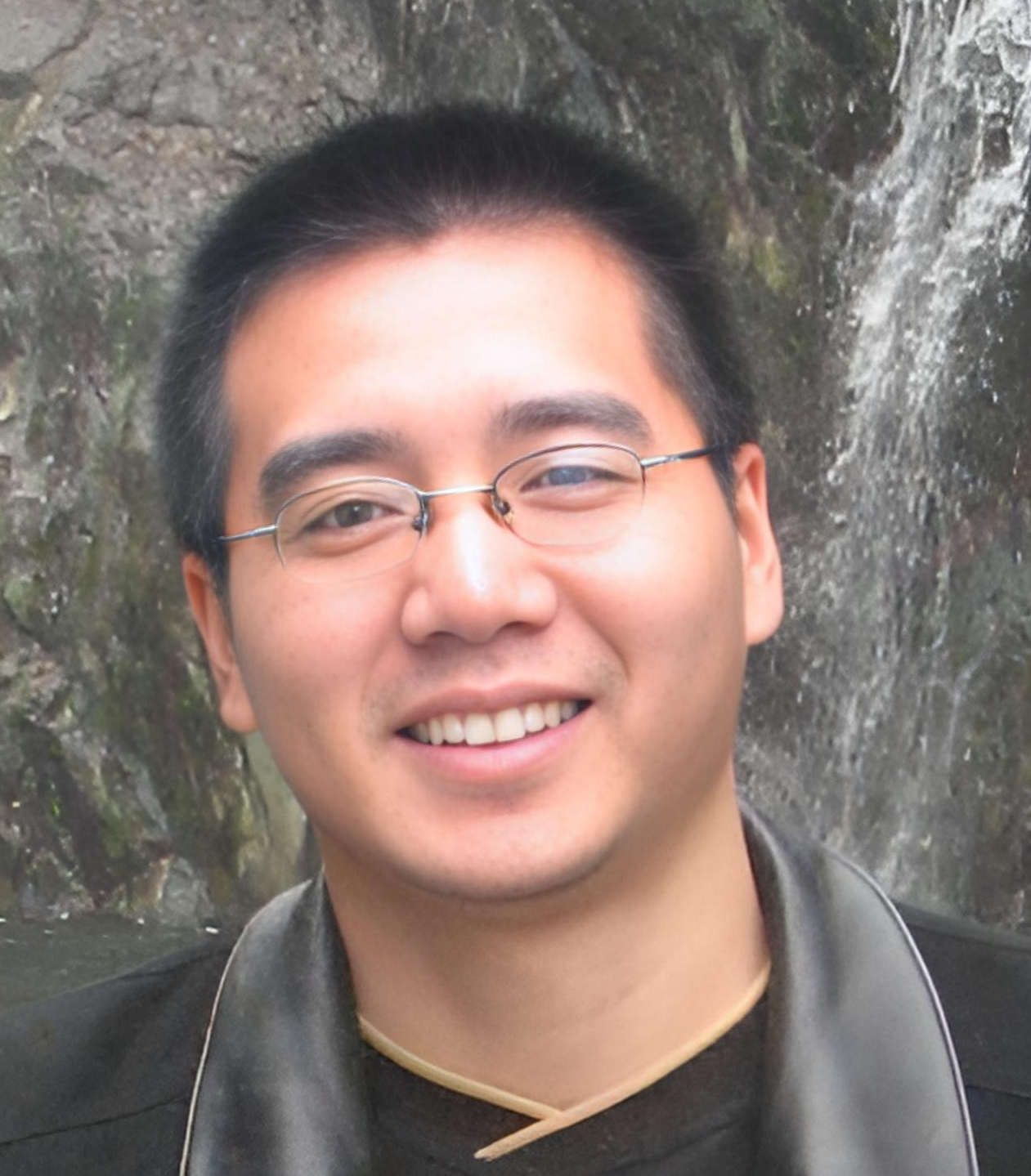}}] 
{Hui Yuan} (S’08–M’12–SM’17) received the B.E. and Ph.D. degrees in telecommunication engineering from Xidian University, Xi’an, China, in 2006 and 2011, respectively. In April 2011, he joined Shandong University, Ji’nan, China, as a Lecturer (April 2011–December 2014), an Associate Professor (January 2015-October 2016), and a Professor (September 2016). From January 2013-December 2014, and November 2017-February 2018, he also worked as a Postdoctoral Fellow (Granted by the Hong Kong Scholar Project) and a Research Fellow, respectively, with the Department of Computer Science, City University of Hong Kong, Hong Kong. From November 2020 to November 2021, he also worked as a Marie Curie Fellow (Granted by the Marie Skłodowska-Curie Individual Fellowships of European Commission) with the Faculty of Computing, Engineering and Media, De Montfort University, United Kingdom. From October 2021 to November 2021, he also worked as a visiting researcher (secondment of the Marie Skłodowska-Curie Individual Fellowships) with the Computer Vision and Graphics group, Fraunhofer Heinrich-Hertz-Institut (HHI), Germany. His current research interests include 3D visual coding, processing, and communication. He served as an Associate Editor for IEEE Transactions on Image Processing (since 2025), and Associate Editor for IEEE Transactions on Consumer Electronics (since 2024), an Associate Editor for IET Image Processing (since 2023), an Area Chair for IEEE ICME (since 2020).
\end{IEEEbiography}
\vspace{-1cm}
\begin{IEEEbiography}
[{\includegraphics[width=1in,height=1.25in,clip,keepaspectratio]{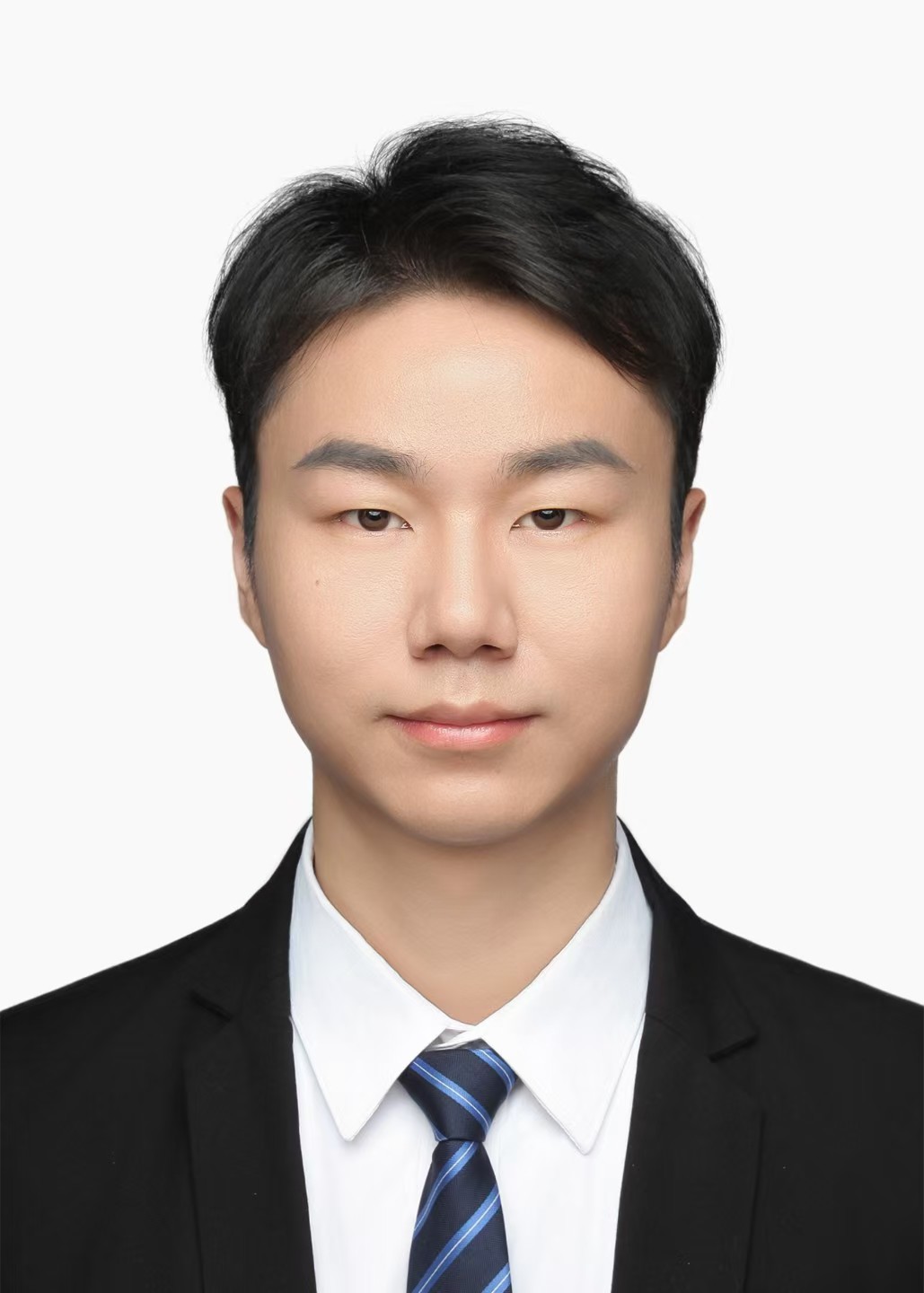}}] 
{Qinglong Miao} received the B.S. degree in mechanical engineering from Shandong University, Jinan, China, in 2020. His research interests include 3D reconstruction and image registration.
\end{IEEEbiography}
\vspace{-4cm}
\begin{IEEEbiography}
[{\includegraphics[width=1in,height=1.25in,clip,keepaspectratio]{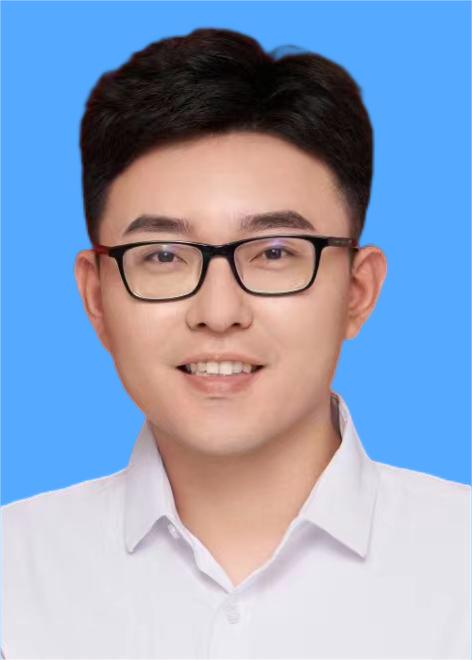}}] 
{Xiaolong Mao} received an M.E.degree from the School of Integrated Circuits, Shandong University, China, in 2021. He is currently pursuing a Ph.D. degree at Shandong University. His research interests include point clouds compression.
\end{IEEEbiography}
\vspace{-4cm}
\begin{IEEEbiography}
[{\includegraphics[width=1in,height=1.25in,clip,keepaspectratio]{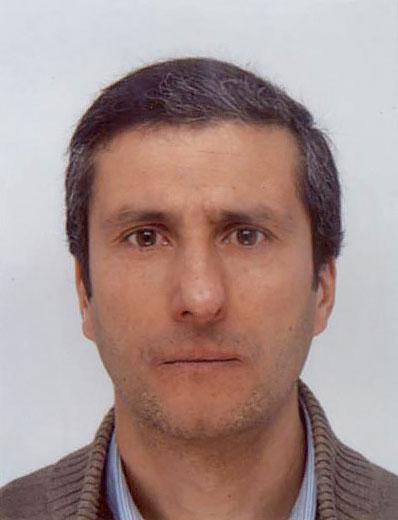}}] 
{Raouf Hamzaoui} (Senior Member, IEEE) received
the M.Sc. degree in mathematics from the University
of Montreal, Montreal, QC, Canada, in 1993, the
Dr.rer.nat. degree from the University of Freiburg,
Freiburg im Breisgau, Germany, in 1997, and the
Habilitation degree in computer science from the
University of Konstanz, Konstanz, Germany, in 2004.
He was an Assistant Professor with the Department
of Computer Science, University of Leipzig, Leipzig,
Germany, and Department of Computer and Information
Science, University of Konstanz. In 2006, he
joined De Montfort University, Leicester, U.K., where he is currently a Professor
of media technology and the Head of the Signal Processing and Communications
Systems Group, Institute of Engineering Sciences. He has authored or
coauthored more than 100 research papers in books, journals, and conferences.
His research has been funded by the EU, DFG, Royal Society, and industry.
He was the recipient of the best paper awards (ICME 2002, PV’07, CONTENT
2010, MESM’2012, and UIC 2019). He was a member of the Editorial Board of
IEEE TRANSACTIONS ON MULTIMEDIA and IEEE TRANSACTIONS ON CIRCUITS
AND SYSTEMS FOR VIDEO TECHNOLOGY.
\end{IEEEbiography}
\vspace{-4cm}
\begin{IEEEbiography}
[{\includegraphics[width=1in,height=1.25in,clip,keepaspectratio]{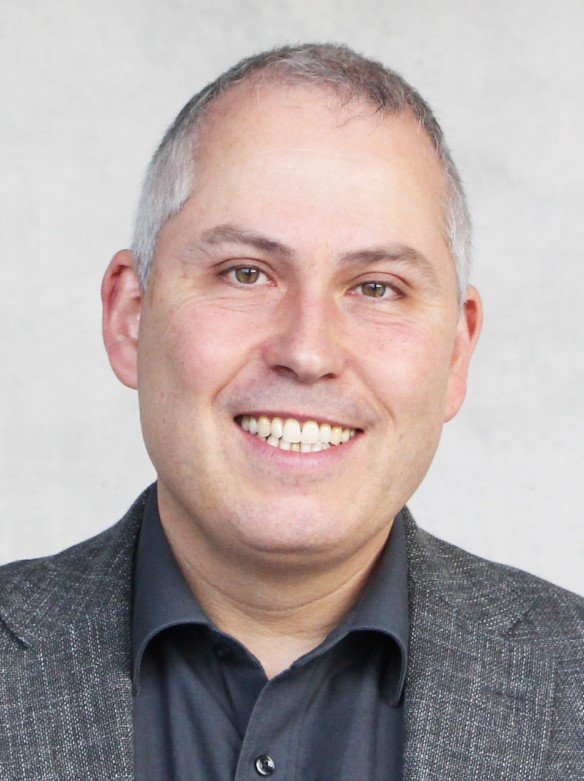}}] 
{Peter Eisert} (Senior Member, IEEE) Peter Eisert is the head of computer vision and graphics at Fraunhofer HHI and the chair on visual computing at Humboldt University. His research interests include 3D scene and surface reconstruction, object tracking, and visualization. Eisert has a PhD in electrical engineering from University of Erlangen Nuremberg. Contact him at peter.eisert@hhi.fraunhofer.de.
\end{IEEEbiography}
\end{document}